# On characterizing Inclusion of Bayesian Networks

Tomáš Kočka
Laboratory for Intelligent Systems
Univ. of Economics Prague
kocka@vse.cz

Remco R. Bouckaert
Crystal Mountain Inf. Tech.
New Zealand
rrb@xm.co.nz

Milan Studený
Inst. Info. Theory and Autom.
Acad. Sci. Czech Rep.
studeny@utia.cas.cz

## Abstract

The *inclusion problem* deals with how to characterize (in graphical terms) whether all independence statements in the model induced by a DAG $K$ are in the model induced by a second DAG $L$. Meek (1997) conjectured that this inclusion holds iff there exists a sequence of DAGs from $L$ to $K$ such that only certain 'legal' arrow reversal and 'legal' arrow adding operations are performed to get the next DAG in the sequence. In this paper we give several characterizations of inclusion of DAG models and verify Meek's conjecture in the case that the DAGs $K$ and $L$ differ in at most one adjacency. As a warming up a rigorous proof of graphical characterizations of equivalence of DAGs is given.

## 1 Introduction

Learning Bayesian network structures requires search in the space of directed acyclic graphs (DAGs). To prove that such learning algorithms return (local) optimal networks, the search space needs to be characterized. A natural way of doing this is to consider the set of conditional independence statements represented by the DAGs in the search space. Once it is known how to characterize all the properties of two DAGs $K$ and $L$ such that independence statements represented in $K$ are represented in $L$ as well, efficient search algorithms can be designed based on this characterization. This characterization problem is called the *inclusion problem*.

Meek (1997) formulated a conjecture which states that inclusion holds iff a special sequence of DAGs $G_1, \ldots, G_n$ starting with $L = G_1$ and ending with $K = G_n$ exists. Here, $G_{i+1}$ is obtained from $G_i$ either by adding an arrow or by performing a single arrow reversal (this arrow reversal is special in that it does not introduce new represented independence statements). Many search algorithms for learning Bayesian networks rely on this conjecture being true for optimality of the learned network structures.

In this paper we give an overview of current state of our research in the inclusion problem. The next section deals with basic concepts and notation, in Section 3 some of our specific concepts are introduced. Section 4 is devoted to equivalence charaterization, Section 5 is an overview of conditions related to the inclusion problem and tries to develop some insight in the nature of the inclusion problem. Section 6 contains the main result: we characterize the case when two DAGs differ in only one adjacency.

## 2 Basic concepts

Throughout the paper the symbol $N$ denotes a non-empty finite set of variables which are identified with nodes of graphs. Juxtaposition $AB$ where $A, B \subseteq N$ will stand for the union $A \cup B$. Independence and dependence statements over $N$ correspond to special *disjoint triplets over* $N$. The symbol $\langle A, B | C \rangle$ denotes a triplet of pairwise disjoint subsets $A, B, C$ of $N$. The symbol $\mathcal{T}(N)$ will denote the class of all disjoint triplets over $N$.

### 2.1 Graphical concepts

A *directed graph* $G$ over a set of *nodes* $N$ is specified by a collection of *arrows*, that is a collection $\mathcal{A}(G)$ of ordered pairs $(u, v)$ of distinct nodes $u, v \in N$, $u \neq v$. We write $u \to v$ in $G$ or $u \to v$ $[G]$ to denote that $(u, v) \in \mathcal{A}(G)$; the symbol of the graph can be omitted if it is clear from the context. In an arrow $u \to v$, denoted alternatively by $v \leftarrow u$, $u$ is called the *tail node* and $v$ the *head node*. Furthermore, we say that $u$ is a *parent* of $v$ and $v$ is a *child* of $u$. The set of parents of $u$ in $G$ will be denoted by $pa_G(u)$, the set of children by $ch_G(u)$. A *subgraph* of a directed graph $G$ over $N$ is determined by a non-empty set of its nodes $A \subseteq N$ and by the set of its arrows which is a subset of $\mathcal{A}(G) \cap (A \times A)$ (strict inclusion is allowed). The *induced subgraph* of $G$ for a non-empty set $B \subseteq N$ is the graph $G_B$ over $B$ having $\mathcal{A}(G_B) = \mathcal{A}(G) \cap (B \times B)$ as the collection of its arrows.

We write $u \leftrightarrow v \ [G]$ to denote that there is an *edge* or an adjacency between distinct nodes $u$ and $v$ in $G$ which means that either $u \to v$ in $G$ or $u \leftarrow v$ in $G$. The set of edges in a directed graph $G$ is the collection $\mathcal{E}(G) = \{ \{u,v\} \ ; \ u \leftrightarrow v \ [G] \}$ of respective two-element subsets of $N$. If there is no edge between $u$ and $v$ in $G$ then we write $u \not\leftrightarrow v \ [G]$ to denote this *non-adjacency*.

A *trail* in $G$ (between nodes $u$ and $v$) is a sequence $\pi$ of (not necessarily distinct) nodes $w_1, \ldots, w_k$, $k \geq 1$ such that $w_i \leftrightarrow w_{i+1} \ [G]$ for every $1 \leq i < k$ (and either $w_1 = u, w_k = v$ or $w_1 = v, w_k = u$). It is called a *path* if all nodes $w_1, \ldots, w_k$ are distinct. A *section* of a path $w_1, \ldots, w_k$, $k \geq 1$ is a path $w_i, \ldots, w_j$ where $1 \leq i \leq j \leq k$. We say that $w_i$, $1 < i < k$ is a *collider node* of a path $\pi$ if $w_{i-1} \to w_i$ in $G$ and $w_i \leftarrow w_{i+1}$ in $G$. Every other node of $\pi$ is called a *non-collider node* of $\pi$. A path $\pi$ in $G$ is called *open* if it has no collider node. We will write $w_1 - w_2 - \ldots - w_k \ [G]$, $k \geq 1$ to denote an open path in $G$.

A trail, resp. a path, is called *directed* if $w_i \to w_{i+1} \ [G]$ for $i = 1, \ldots, k-1$. We say that it is a path from a node $u$ to a node $v$ (from $A \subseteq N$ to $B \subseteq N$) if $w_1 = u$ and $w_k = v$ ($w_1 \in A$ and $w_k \in B$). A node $u$ is called an *ancestor* of a node $v$ in $G$ (alternatively $v$ is a *descendant* of $u$ in $G$) if there is a directed path from $u$ to $v$ in $G$. Observe that every node is its own ancestor and its own descendant since paths with only a single node are regarded as directed paths. The symbol $an_G(A)$ will denote the set of all ancestors of nodes of a set $A \subseteq N$ in $G$ and $ds_G(u)$ the set of descendants of a node $u$ in $G$.

A *directed cycle* is a directed trail $w_1, \ldots, w_k$, $k \geq 3$ such that $w_1 = w_k$ and $w_1, \ldots, w_{k-1}$ are distinct nodes. A *directed acyclic graph* (DAG or ADG) is a directed graph without directed cycles. Note that every trail (path) in a DAG, which has been defined as a sequence of nodes, has uniquely determined the (type of) arrows connecting consecutive nodes and therefore $|\mathcal{A}(G)| = |\mathcal{E}(G)|$ for every DAG $G$. Another observation is that a subgraph of a DAG is also a DAG. A well-known equivalent definition of a DAG is as follows: $G$ is a directed graph and all its nodes can be ordered into a sequence $u_1, \ldots, u_n$, $n \geq 1$ such that $pa_G(u_i) \subseteq \{u_j \ ; 1 \leq j < i\}$ for every $i = 1, \ldots, n$. An ordering of this type is called a *causal ordering* for $G$. A *terminal node* is a node without children.

Well-known fact is that every DAG has at least one terminal node. We say that distinct nodes $u, v, w$ form an *immorality* in a directed graph $G$ and write $(u, v) \rightsquigarrow w \ [G]$ if $u \to w$ in $G$, $v \to w$ in $G$ and $u \not\leftrightarrow v \ [G]$. In fact, an immorality in a DAG $G$ is nothing but a special induced subgraph of $G$.

An *undirected graph* $H$ over $N$ is specified by a collection $\mathcal{L}(H)$ of two-element subsets of $N$ which are called *lines* in $H$. By the *underlying graph* of a directed graph $G$ over $N$ is understood an undirected graph $H$ for which $\mathcal{L}(H) = \mathcal{E}(G)$.

### 2.2 Induced models

One of possible ways of associating independence models with DAGs is by d-separation criterion from (Pearl 1988). Let $\pi : w_1, \ldots, w_k$, $k \geq 1$ be a path in a DAG $G$. The path $\pi$ is called *active* with respect to a set $C \subseteq N$ (shortly w.r.t. $C$) if

- every non-collider node of $\pi$ is not in $C$,
- every collider node of $\pi$ has a descendant in $C$.

Suppose that $\langle A, B | C \rangle \in \mathcal{T}(N)$ is a disjoint triplet over $N$; one says that $A$ and $B$ are *d-connected* given $C$ in a DAG $G$, written $A \top\!\!\!\top B \,|\, C \ [G]$, if there exists a path between a node $a \in A$ and a node $b \in B$ in $G$ which is active w.r.t. $C$. In the opposite case one says that $A$ and $B$ are *d-separated* by $C$ in $G$ which is denoted by $A \perp\!\!\!\perp B \,|\, C \ [G]$. We also say that $\langle A, B | C \rangle$ is *represented* in $G$ according to the d-separation criterion. The *induced independence model* $\mathcal{I}(G)$ and the induced dependence model $\mathcal{D}(G)$ are as follows:

$$\mathcal{I}(G) = \{ \langle A, B | C \rangle \in \mathcal{T}(N) \,; \quad A \perp\!\!\!\perp B \,|\, C \ [G] \},$$

$$\mathcal{D}(G) = \{ \langle A, B | C \rangle \in \mathcal{T}(N) \,; \quad A \top\!\!\!\top B \,|\, C \ [G] \}.$$

Note that an alternative to the d-separation criterion is the moralization criterion introduced by Lauritzen et. al. (1990).

## 3 Specific concepts

This section describes some specific concepts we regards as relevant to the inclusion problem. Our concept of dependence complex has, despite its long technical definition, a good intuitive sense and hopefully brings insight into the problem. For example, it is essential for our later Conjecture 2.

### 3.1 Dependence complex

Let $G$ be a DAG over $N$, $C \subseteq N$ and $a, b \in N \setminus C$ are distinct nodes. Let $\pi : w_1, \ldots, w_k$, $k \geq 2$ be a path in $G$ between $a = w_1$ and $b = w_k$ which is active w.r.t. $C$. Every collider node $d$ of $\pi$ which is not in $C$ has necessarily a descendant $c \in C$, $c \neq d$ in $G$. By a *rope* for $d$ (with respect to $\pi$) will be understood a directed path $\rho : t_1, \ldots, t_r$, $r \geq 2$ in $G$ from $d = t_1$ to a node $c = t_r$ in $C$ such that

- $\rho$ is outside $C$ with exception of $c$, i.e. $t_1, \ldots, t_{r-1} \notin C$,

- $\rho$ does not share a node with $\pi$ except $d$, i.e. $t_2, \ldots, t_r \notin \{w_1, \ldots, w_k\}$.

Let us denote by $col(\pi, C)$ the set of collider nodes of $\pi$ which are outside $C$.

A *dependence complex* (between $a$ and $b$) for $C$ in $G$ is a special subgraph $\kappa$ of $G$. First, we specify the collection of arrows of a dependence complex. Each complex $\kappa$ (for $C$) is specified by the following items:

- a path $\pi$ in $G$ which is active w.r.t. $C$,

- a collection of ropes $\{\rho(d)\,;\, d \in col(\pi, C)\,\}$ with respect to $\pi$,

where every collider node $d \in col(\pi, C)$ has assigned only one rope $\rho(d)$ in $\kappa$ and the ropes for distinct collider nodes do not share a node. The collection of arrows in $\kappa$ then consists of the arrows involved in $\pi$ and in $\rho(d)$ for $d \in col(\pi, C)$. Second, we specify the set of nodes of a dependence complex as the set of head nodes and tail nodes of the chosen arrows. Thus, $\kappa$ is a subgraph of $G$ which need not have the whole set $N$ as the set of nodes. Instead of dependence complex for $C$ in $G$ we say shortly $C$-*complex* in $G$ (between $A$ and $B$ in case $a \in A$ and $b \in B$).

Let us emphasize that every dependence complex $\kappa$ uniquely decomposes into the path $\pi$ and the collection of ropes. Indeed, every node of a given subgraph $\kappa$ of $G$ (which was constructed as a dependence complex in $G$ for a set $C \subseteq N$ and $a, b \in N \setminus C$) can be classified into one of three groups according to the number of edges of $\kappa$ 'entering' the node (this number varies from 1 to 3). The conditions required in the definition of a dependence complex above imply that a node of $\kappa$ has 3 'entering' edges iff it belongs to $col(\pi, C)$. Moreover, a node of this kind is twice a head node and once a tail node: this determines which of the 'branches' outgoing the node is a rope.

LEMMA 3.1 Let $G$ be a DAG over $N$, $C \subseteq N$ and $a, b \in N \setminus C$ are distinct nodes. Then $a \mathbin{\top\!\!\!\top} b \,|\, C \,[G]$ iff there exists a dependence complex in $G$ between $a$ and $b$ for $C$.

**Proof:** A complete proof can be found in (Kočka et. al. 2001). The main idea of the necessity proof is to choose an active path $\pi$ with respect to $C$ with minimal number of collider nodes and to choose for every $d \in col(\pi, C)$ as a rope a directed path from $d$ to $C$ with minimal number of arrows. □

Note that the concept of dependence complex corresponds to the concept 'path-with-tails' mentioned by Matúš (1997).

### 3.2 Composite dependence statements

The point is that every dependence complex ensures validity of a certain composite dependence statement. Given a DAG $G$ over $N$, distinct nodes $u, v \in N$ and disjoint sets $S, T \subseteq N \setminus \{u, v\}$ we interpret the symbol $u \mathbin{\top\!\!\!\top} v \,|\, + T - S \,[G]$ as the condition

$$u \mathbin{\top\!\!\!\top} v \,|\, W \,[G] \quad \text{whenever } T \subseteq W \subseteq N \setminus \{u, v\} \cup S.$$

In words, $u$ and $v$ are (conditionally) dependent in $G$ given any superset of $T$ which is disjoint with $S$. In case that $T$ respectively $S$ is empty the symbols $+T$ respectively $-S$ are omitted; if both $T$ and $S$ is empty we write $\star$ instead of $+T - S$. Observe that if $\kappa$ is a dependence complex between $a$ and $b$ for $C$ in $G$, $S$ is the set of non-collider nodes of the respective active path $\pi$ except $a, b$ and $T$ is the set of nodes of $\kappa$ belonging to $C$ then $a \mathbin{\top\!\!\!\top} b \,|\, + T - S \,[G]$. Thus, every dependence complex ensures validity of a composite dependence statement but the converse in not true in general.

We give a certain graphical characterization of some composite dependence statements of this kind below. These auxiliary results were proved in (Studený 1997) as Lemmas 3.1, 3.2 and 3.3 in wider context of chain graphs; see also (Verma Pearl 1991).

LEMMA 3.2 Let $G$ be a DAG over $N$ and $u, v \in N$ are distinct nodes. Then

$$u \perp\!\!\!\perp v \,|\, pa_G(u) pa_G(v) \,[G] \quad \text{whenever } u \not\leftrightarrow v \,[G].$$

LEMMA 3.3 Let $G$ be a DAG over $N$ and $u, v \in N$ are distinct nodes. Then

$$u \leftrightarrow v \,[G] \quad \text{iff} \quad u \mathbin{\top\!\!\!\top} v \,|\, \star \,[G].$$

LEMMA 3.4 Let $G$ be a DAG over $N$ and $u, v, w \in N$ are distinct nodes such that $u \leftrightarrow w \,[G]$, $v \leftrightarrow w \,[G]$ and $u \not\leftrightarrow v \,[G]$. Then

$$(u, v) \rightsquigarrow w \,[G] \quad \text{iff} \quad u \mathbin{\top\!\!\!\top} v \,|\, + w \,[G].$$

Note that pure composite dependence statement $u \mathbin{\top\!\!\!\top} v \,|\, + w \,[G]$ can be characterized in graphical terms as follows (Kočka et. al. 2001): either $u \leftrightarrow v \,[G]$ or $u \rightarrow t \leftarrow v$ in $G$ and $t \in an_G(w)$ for some $t \in N$.

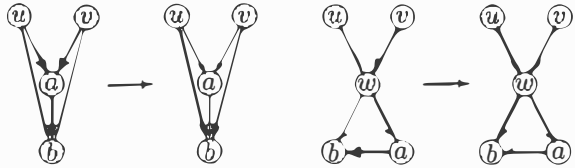

Figure 1: Rope modification (shortening) in $L$.

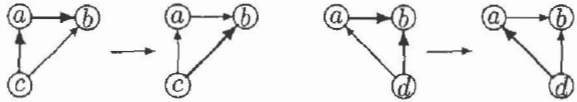

Figure 2: Path shortening in $L$.

## 4 Equivalence of DAGs

In this section we deal with a well understood special case of the inclusion problem - the equivalence problem. It is the problem how to recognize whether two given DAGs $K$ and $L$ over $N$ induce the same independence model. It is of special importance to have an easy rule how to recognize that two DAGs are equivalent in this sense and an easy way to get from $L$ to $K$ in terms of some elementary operations on graphs. These issues were already treated by Verma and Pearl (1991), Heckerman et. al. (1994), Chickering (1995) and Frydenberg (1990) in the context of chain graphs.

By a *legal arrow reversal* is understood the change of a DAG $L$ into a directed graph $K$ by replacement of an arrow $a \to b$ (in $L$) by $b \to a$ (in $K$) under the condition that $pa_L(a) \cup a = pa_L(b)$ (here $a, b \in N$ are some distinct nodes).

Note that Chickering (1995) used *covered edge* and Meek (1997) *covered arc* instead. The following observations follow from Lemma 1 in (Chickering 1995).

OBSERVATION 4.1 The result of a legal arrow reversal operation is a DAG.

LEMMA 4.1 Let $K$ and $L$ be DAGs over $N$ such that $K$ is obtained from $L$ by a legal arrow reversal. Then $\mathcal{I}(K) = \mathcal{I}(L)$.

An alternative proof of Lemma 4.1 which uses the concept of dependence complex can be found in (Kočka et. al. 2001). Basic idea is to apply Lemma 3.1 to $a \perp\!\!\!\perp b \mid C$ and show that every $C$-complex between $a$ and $b$ in $L$ with minimal number of edges must be in $K$ (and conversely). To prove this fact by contradiction modifications (shortening) of the considered complex in $L$ indicated by Figures 1 and 2 are made.

LEMMA 4.2 Supposing $K$ and $L$ are DAGs over $N$ the following three conditions are equivalent:

(1) $\mathcal{I}(K) = \mathcal{I}(L)$,

(2) $\mathcal{E}(K) = \mathcal{E}(L)$ and the graphs $K$ and $L$ have the same immoralities,

(3) there exists a sequence $G_1, \ldots, G_m$, $m \geq 1$ of DAGs over $N$ such that $G_1 = L, G_m = K$ and $G_{i+1}$ is obtained from $G_i$ by legal arrow reversal for $i = 1, \ldots, m-1$.

Note that the equivalence (1) $\Leftrightarrow$ (2) was proved in (Verma Pearl 1991), in the framework of chain graphs in (Frydenberg 1990); the equivalence (1) $\Leftrightarrow$ (3) was proved in (Heckerman et. al. 1994) and (Chickering 1995). Our proof is different in that it is constructive and provides an algorithm for finding the sequence mentioned in (3). The algorithm can be applied in case of the inclusion problem - see Section 6.

**Proof:** We show (1) $\Rightarrow$ (2) $\Rightarrow$ (3) $\Rightarrow$ (1). The implication (1) $\Rightarrow$ (2) is an easy consequence of Lemmas 3.3 and 3.4 as $\mathcal{I}(K) = \mathcal{I}(L)$ is equivalent to $\mathcal{D}(K) = \mathcal{D}(L)$.

The proof of (2) $\Rightarrow$ (3) is done by induction on $|N|$. The induction hypothesis for $n \geq 1$ is that (2) $\Rightarrow$ (3) holds for any pair of DAGs $K, L$ over $N$ with $|N| \leq n$. This is evident for $n = 1$. Assume $n = |N| \geq 2$ and that the implication holds for DAGs over $N'$ with $|N'| < n$. The first step is to choose a terminal node $t \in N$ in $K$ and put $P = pa_L(t), C = ch_L(t)$. Observe that $\mathcal{E}(K) = \mathcal{E}(L)$ implies $pa_K(t) = P \cup C$. One can distinguish two cases

**I.** $C = \emptyset$ which means $pa_L(t) = pa_K(t)$,

**II.** $C \neq \emptyset$ which means $pa_K(t) \setminus pa_L(t) \neq \emptyset$.

If $C = \emptyset$ then introduce $L'$ respectively $K'$ as the induced subgraph of $L$ respectively $K$ for $N' \equiv N \setminus \{t\}$. By the induction hypothesis a desired sequence of $L' = G'_1, \ldots, G'_m = K'$, $m \geq 1$ exists. Introduce $G_i$ as the graph over $N$ obtained from $G'_i$ by adding a bunch of arrows from nodes of $P$ to $t$ for $i = 1, \ldots, m$. It is easily seen that $G_{i+1}$ is obtained from $G_i$ by a legal arrow reversal for $i = 1, \ldots, m-1$.

If $C \neq \emptyset$ then choose $c \in C$ such that no other $c' \in C$ is an ancestor of $c$ in $L$. This choice is always possible and ensures that $pa_L(c) \cap C = \emptyset$. The second step is to observe $P \subseteq pa_L(c)$. Indeed, suppose that $p \not\to c \, [L]$ for some $p \in P$. Then, $p \not\to c \, [K]$, $p \leftrightarrow t \, [K]$ and $c \leftrightarrow t \, [K]$ by $\mathcal{E}(K) = \mathcal{E}(L)$. Since $t$ is a terminal node in $K$ one has $(p, c) \rightsquigarrow t \, [K]$ and $(p, c) \rightsquigarrow t \, [L]$ by (2). This however contradicts the fact $t \to c$ in $L$. Thus, necessarily $p \leftrightarrow c \, [L]$. Since $L$ is acyclic and $p \to t \to c$ in $L$ it implies $p \to c$ in $L$. The third observation is that $pa_L(c) \subseteq P \cup \{t\}$. Indeed, suppose for contradiction that there exists $y \in N \setminus P$, $y \neq t$ such that $y \to c$ in $L$ (see Figure 3 for illustration where, however, arrows from $P$ to $C$ are omitted for sake of lucidity). Since $y \notin P$ and $y \notin C$ (because of the choice of $c$) one has $t \not\leftrightarrow y$ $[L]$. Thus $y \to c \leftarrow t$ in $L$ implies $(y, t) \rightsquigarrow c$ $[L]$ and $(y, t) \rightsquigarrow c$ $[K]$ by (2). This contradict the fact $c \to t$ in $K$. Therefore, necessarily $pa_L(c) = P \cup \{t\}$.

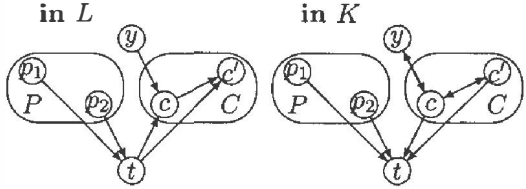

Figure 3: Proof of $pa_L(c) \subseteq P \cup \{t\}$ by contradiction.

The fact $pa_L(c) = pa_L(t) \cup \{t\}$ means that the arrow $t \to c$ in $L$ can be legally reversed. By Lemma 4.1 and (1)$\Rightarrow$(2) the same procedure can be repeated until all arrows in $C$ are legally reversed. Thus, a sequence $L = G_1, \ldots, G_k$, $k \geq 2$ is constructed by legal arrow reversals such that $t$ has the same parents in $G_k$ as in $K$. Then, the case **I.** occurs for the pair $(G_k, K)$ which was already solved. This concludes the induction step.

The proof of (3) $\Rightarrow$ (1) can be done by repetitive application of Lemma 4.1. □

## 5　Conditions for inclusion

In this section, some characterizations of inclusion are given in terms of graphical conditions and insight is obtained on the nature of such conditions. We give an overview of various necessary conditions on DAGs $K$ and $L$ over $N$ for validity of inclusion $\mathcal{I}(K) \subseteq \mathcal{I}(L)$. Everybody who takes up the inclusion problem finds almost immediately that the following three *basic conditions* are necessary for inclusion $\mathcal{I}(K) \subseteq \mathcal{I}(L)$:

(a) $u \leftrightarrow v$ $[L]$ $\Rightarrow$ $u \leftrightarrow v$ $[K]$,

(b̃) $(u, v) \rightsquigarrow w$ $[L]$ $\Rightarrow$ $u \leftrightarrow v$ $[K]$ or $(u, v) \rightsquigarrow w$ $[K]$,

(c̃) $(u, v) \rightsquigarrow w$ $[K]$ $\Rightarrow$
　　$u \not\leftrightarrow w$ $[L]$ or $w \not\leftrightarrow v$ $[L]$ or $(u, v) \rightsquigarrow w$ $[L]$,

(see Observation 5.1 and 5.2 below). Note that the condition (b̃) respectively (c̃) can be under (a) equivalently formulated as follows (observe that $X \Rightarrow Y$ is equivalent to $\neg Y \Rightarrow \neg X$):

(b) $u \to w \leftarrow v$ $[L]$ $\Rightarrow$ $u \leftrightarrow v$ $[K]$ or $u \to w \leftarrow v$ $[K]$,

(c) $u - w - v$ $[L]$ $\Rightarrow$ $u \leftrightarrow v$ $[K]$ or $u - w - v$ $[K]$.

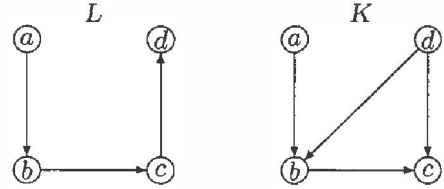

Figure 4: Basic conditions are not sufficient.

The conditions are also sufficient in the following rather special case where two DAGs have the same number of edges.

**LEMMA 5.1** Suppose that $K, L$ are DAGs over $N$ such that $|\mathcal{E}(K)| \leq |\mathcal{E}(L)|$. Then the conditions (a), (b̃) and (c̃) are necessary and sufficient for $\mathcal{I}(K) \subseteq \mathcal{I}(L)$.

**Proof:** This follows from Lemma 4.2. The condition (a) says $\mathcal{E}(L) \subseteq \mathcal{E}(K)$ which together with $|\mathcal{E}(K)| \leq |\mathcal{E}(L)|$ implies $\mathcal{E}(K) = \mathcal{E}(L)$. The conditions (b̃) and (c̃) then imply that $K$ and $L$ have the same immoralities. □

Verma and Pearl (1988) formulated (using another notation) in one of their technical reports three necessary conditions on DAGs $K$ and $L$ over $N$. We call them *Verma's conditions*.

(i) $u \leftrightarrow v$ $[L]$ $\Rightarrow$ $u \leftrightarrow v$ $[K]$,

(ii) $u \top\!\!\top v \mid +w$ $[L]$ $\Rightarrow$ $u \top\!\!\top v \mid +w$ $[K]$,

(iii) $u \top\!\!\top v \mid +w$ $[K]$, $u \leftrightarrow w$ $[L]$, $w \leftrightarrow v$ $[L]$ $\Rightarrow$
　　　$u \top\!\!\top w \mid +w$ $[L]$ or $u \leftrightarrow v$ $[K]$.

Note that (i) is nothing but (a) and one can show that (iii) is under (a) equivalent to (c̃) by Lemma 3.4. However neither the basic conditions nor Verma's conditions are sufficient for $\mathcal{I}(K) \subseteq \mathcal{I}(L)$ in general as the example in Figure 4 shows. Condition (a) evidently holds since every edge in $L$ is in $K$. Condition (b̃) holds since no configuration $(u, v) \rightsquigarrow w$ exists in $L$ and condition (c̃) applies to $(a, d) \rightsquigarrow b$ for which the edge $b \leftrightarrow d$ is missing in $L$. So, the conditions (a), (b̃) and (c̃) are evidently fulfilled in that case but one has $a \perp\!\!\!\perp d \mid \emptyset$ $[K]$ while $a \top\!\!\top d \mid \emptyset$ $[L]$ which implies $\neg\{\mathcal{I}(K) \subseteq \mathcal{I}(L)\}$.

Let $K$ and $L$ are DAGs over $N$. We will call the following 3 conditions the *inclusion conditions* for $K$ in $L$ (here, $u, v, w$ are distinct elements of $N$):

(a) $u \leftrightarrow v$ $[L]$ $\Rightarrow$ $u \leftrightarrow v$ $[K]$,

(b̃) $(u, v) \rightsquigarrow w$ $[L]$ $\Rightarrow$ $u \leftrightarrow v$ $[K]$ or $(u, v) \rightsquigarrow w$ $[K]$,

(∗) $(u, v) \rightsquigarrow w$ $[K]$ $\Rightarrow$ $u \perp\!\!\!\perp v \mid pa_K(u) pa_K(v)$ $[L]$.

Clearly, Lemmas 3.3, 3.4 and 3.2 imply almost immediately the following observation (for $(\tilde{b})$ use (a)).

OBSERVATION 5.1 *The inclusion conditions for $K$ in $L$ are necessary for validity of $\mathcal{I}(K) \subseteq \mathcal{I}(L)$.*

OBSERVATION 5.2 *The inclusion conditions for $K$ in $L$ imply the basic necessary conditions (a), $(\tilde{b})$, $(\tilde{c})$.*

**Proof:** It suffices to verify $(\tilde{c})$. If $(u,v) \rightsquigarrow w\ [K]$ then let $W \equiv pa_K(u)pa_K(v)$ and observe $w \notin W$. Suppose that the conclusion of $(\tilde{c})$ is not valid. This means either $u \leftrightarrow v\ [L]$ which contradicts the fact $u \not\leftrightarrow v\ [K]$ by (a), or $u \not\leftrightarrow v\ [L]$ and the path $u,w,v$ in $L$ has no collider nodes. This path is then active w.r.t. $W$ (as $w \notin W$) which means $u \not\perp v | W\ [L]$. However, the condition $(*)$ implies $u \perp\!\!\!\perp v | W\ [L]$ which contradicts that fact. Thus, the conclusion of $(\tilde{c})$ must hold. □

CONJECTURE 1 *The inclusion conditions for $K$ in $L$ are sufficient for $\mathcal{I}(K) \subseteq \mathcal{I}(L)$.*

The inclusion conditions can be strengthened to get one necessary and sufficient condition called the *enforced inclusion condition*:

$(**)\ u \not\leftrightarrow v\ [K]\ \Rightarrow\ u \perp\!\!\!\perp v | pa_K(u)pa_K(v)\ [L].$

LEMMA 5.2 *Let $K$ and $L$ be DAGs over $N$. Then $\mathcal{I}(K) \subseteq \mathcal{I}(L)$ iff the enforced inclusion condition $(**)$ holds.*

**Proof:** If $\mathcal{I}(K) \subseteq \mathcal{I}(L)$ then $(**)$ by Lemma 3.2. A well-known result from (Verma Pearl 1990) implies that to show $(**) \Rightarrow \{\mathcal{I}(K) \subseteq \mathcal{I}(L)\}$ it suffices to verify $\mathcal{L}_{K,\theta} \subseteq \mathcal{I}(L)$ for an input list generated by a causal ordering $\theta : u_1, \ldots, u_n$ for $K$. It can be shown that $\langle u_i, u_1 \ldots u_{i-1} \setminus pa_K(u_i) | pa_K(u_i) \rangle \in \mathcal{I}(L)$ by induction on $i = 1, \ldots, n$. The essential tool for proving this is the observation that $\mathcal{I}(L)$ is a graphoid which satisfies the composition property from (Pearl 1988). The details of the proof can be found in (Kočka et. al. 2001). □

Some graphical conditions mentioned above were local in the sense that their verification depends on subgraphs involving only a few nodes. Now, we show that one cannot expect full characterization of $\mathcal{I}(K) \subseteq \mathcal{I}(L)$ in terms of conditions of this type. Consider the independence model whose only non-trivial independence statement corresponds to a disjoint triplet $\langle a, b | Z \rangle$. Figure 5 shows a DAG $K$ of that represents a model of this type. It is very easy to construct a DAG $L$ such that there is only a single path from $a$ to $b$ in $L$ which ensures $a \perp\!\!\!\perp b | Z\ [L]$. Obviously, a path of this type can be made as long and complex as one likes which means that plenty of these DAGs $L$ exists.

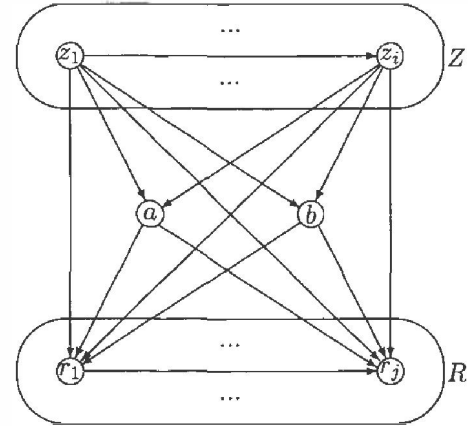

Figure 5: A counterexample to locality of conditions.

The key insight here is that this general example shows that one has to look for a set of conditions in which at least one has a *non-local* aspect.

In the case when $K$ and $L$ differ in at most one adjacency, the following set of local *graphical conditions* characterize inclusion.

(a) $u \leftrightarrow v\ [L] \Rightarrow u \leftrightarrow v\ [K]$,

(b) $u \rightarrow w \leftarrow v\ [L] \Rightarrow u \leftrightarrow v\ [K]$ or $u \rightarrow w \leftarrow v\ [K]$,

(c) $u - w - v\ [L] \Rightarrow u \leftrightarrow v\ [K]$ or $u - w - v\ [K]$,

(d) $u \rightarrow w \leftarrow t \leftrightarrow v\ [L] \Rightarrow u \leftrightarrow v\ [K]$ or $u - t - v\ [K]$ or $u \rightarrow w \leftarrow v\ [K]$ or $u \rightarrow w \leftarrow t \leftrightarrow v\ [K]$,

(e) $u - w - t - v\ [L] \Rightarrow u \leftrightarrow v\ [K]$ or $u - w - v\ [K]$ or $u - t - v\ [K]$ or $u - w - t - v\ [K]$.

LEMMA 5.3 *The conditions (a)-(e) are implied by the inclusion conditions for $K$ in $L$. In particular, they are necessary for $\mathcal{I}(K) \subseteq \mathcal{I}(L)$. Moreover, they remain valid if $K$ respectively $L$ is replaced by an equivalent graph.*

**Proof:** The proof is in (Kočka et. al. 2001). The invariance relative to equivalence can be shown by reformulating these conditions in terms of respective minimal dependence complexes, i.e. complexes without proper subcomplexes, which appear to be invariants of equivalence classes of DAGs. □

Later Lemma 6.1 implies that the conditions (a)-(e) are also sufficient in case $|\mathcal{E}(K)| \leq |\mathcal{E}(L)| + 1$. Note that $u \leftrightarrow v$ ensures $u \not\perp v | \star$, $u \rightarrow w \leftarrow v$ ensures $u \not\perp v | + w$, $u - w - v$ ensures $u \not\perp v | - w$, $u \rightarrow w \leftarrow t \leftrightarrow v$ ensures $u \not\perp v | +w-t$ and $u-w-t-v$ ensures $u \not\perp v | - wt$. Thus, (a)-(e) can be intuitively interpreted as follows. If $L$ has a dependence complex which ensures the validity of a certain composite dependence statement then $K$ has a 'subcomplex'

which also ensures the validity of that composite dependence statement. We think that the idea behind the construction of these conditions can be extended to a general case and dare to formulate the following conjecture.

CONJECTURE 2 The following condition

(D) Every (minimal) dependence complex in $L$ has a (minimal) subcomplex in $K$.

is necessary and sufficient for $\mathcal{I}(K) \subseteq \mathcal{I}(L)$.

However, one has to specify carefully and formally when a complex in $K$ is a subcomplex of a given complex in $L$. This involves a lot of technicalities - an attempt is made in (Kočka et. al. 2001).

## 6 Meek's conjecture

In this section Meek's conjecture (1997) is recalled and verified in a special case when DAGs differ in at most one adjacency.

By *legal arrow adding* is understood the change of a DAG $L$ into a directed graph $K$ by adding an arrow $a \to b$ in $K$ which is not in $L$ such that the resulting graph $K$ is a DAG.

The following observation is evident.

OBSERVATION 6.1 If $K$ is obtained from $L$ by legal arrow adding then $\mathcal{I}(K) \subseteq \mathcal{I}(L)$.

CONJECTURE 3 (Meek 1997)
The condition $\mathcal{I}(K) \subseteq \mathcal{I}(L)$ is equivalent to the existence of a sequence of DAGs $G_1, \ldots, G_n$, $n \geq 1$ such that $L = G_1$, $K = G_n$ and $G_{i+1}$ is obtained from $G_i$ by applying either the operation legal arrow reversal or the operation of legal arrow adding for $i = 1, \ldots, n-1$.

Observation 6.1 and Lemma 4.1 imply that the existence of above sequence implies $\mathcal{I}(K) \subseteq \mathcal{I}(L)$. Currently there is no known counterexample against the Meek's conjecture.

REMARK 6.1 One may think that a simpler version of Meek's conjecture could be valid. Namely that for two DAGs $K$ and $L$ over $N$ the inclusion $\mathcal{I}(K) \subseteq \mathcal{I}(L)$ implies that there exists a sequence of DAGs $L, \ldots, L_*, \ldots, K_*, \ldots, K$ where $L_*$ is obtained from $L$ by a sequence of legal arrow reversals, $K_*$ is obtained from $L_*$ by a sequence of legal arrow addings and $K$ is obtained from $K_*$ by a sequence of legal arrow reversals. This is to warn the reader that this is not the truth. A counterexample is shown in Figure 6. The example shows two DAGs $K$ and $L$ such that there

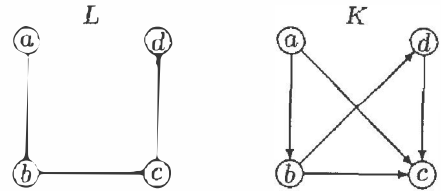

Figure 6: Meek's conjecture cannot be simplified.

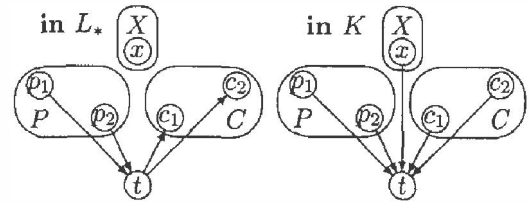

Figure 7: General starting situation.

are no equivalent DAGs $K_*$ and $L_*$ which have the same terminal node ($c$ is always a terminal node in $K_*$ but not in $L_*$). Thus, it is not possible to obtain any $K_*$ from any $L_*$ by legal arrow addings as causal orderings of these always differ. On the other hand, $\mathcal{I}(K) \subseteq \mathcal{I}(L)$ since $K$ can be obtained from $L$ by adding of $b \to d$, then reversal of $c \to d$ and adding of $a \to c$. ◊

LEMMA 6.1 Let $K, L$ are DAGs over $N$ satisfying the conditions (a)-(e) and the condition

(•) $|\mathcal{E}(K)| = |\mathcal{E}(L)| + 1$.

Then there exists a sequence $G_1, \ldots, G_n$, $n \geq 2$ of DAGs over $N$ and $1 \leq m < n$ such that

- $G_1 = L$, $G_{i+1}$ is obtained from $G_i$ by legal arrow reversal for $i = 1, \ldots, m-1$,

- $G_{m+1} \equiv K_*$ is obtained from $G_m \equiv L_*$ by legal arrow adding,

- $G_{i+1}$ is obtained from $G_i$ by legal arrow reversal for $i = m+1, \ldots, n-1$, $G_n = K$.

**Proof:** This is only a sketch of the proof; a complete proof can be found in (Kočka et. al. 2001). It is done by induction on the number of vertices $|N|$. Assume that the statement of the lemma is valid for any pair of DAGs over a set of variables $N'$ with $|N'| < |N|$.

The first step to verify its validity for $N$ is to choose a terminal node $t$ in $K$. It may happen that $t \to y$ in $L$ for some $y \in N$. The second step is to perform legal arrow reversals of these arrows as long as this is possible. Thus, a sequence $L, \ldots, L_*$ of DAGs over $N$ is created by legal arrow reversals. Put

$$P = pa_{L_*}(t),\ C = ch_{L_*}(t),\ X = pa_K(t) \setminus (P \cup C).$$

The situation is depicted in Figure 7. By Lemma 4.2 $L$ and $L_*$ are equivalent which means they have the same underlying graph and immoralities. Since no arrow $t \to y$ in $L_*$ can be legally reversed at least one of the following four cases must occur.

I. $C = \emptyset = X$,

II. $C = \emptyset$ and $X \neq \emptyset$,

III. $P \setminus pa_{L_*}(c) \neq \emptyset$ for some $c \in C$,

IV. $pa_{L_*}(c) \setminus P \cup \{t\} \neq \emptyset$ for some $c \in C$.

In case **I.** the induction hypothesis is applied to the induced subgraphs of $L_*$ and $K$ for $N \setminus \{t\}$ (cf. the proof of Lemma 4.2).

In the other cases a suitable arrow is added to $L_*$ and the resulting graph $K_*$ is shown to be a DAG equivalent to $K$ (with help of the condition (2) of Lemma 4.2). This is done by showing that all new immoralities created in $K_*$ are in $K$ as well and that every immorality in $K$ occurs in $K_*$ owing to the choice of added arrow. The arguments are based on the conditions (a)-(e) only.

Which arrow is added depends on the case which occurs. It is $x \to t$ for $x \in X$ in case **II.**, $p \to c$ where $c \in C, p \in P \setminus pa_{L_*}(c)$ in case **III.** and an arrow $x \to t$ where for suitable $c \in C$ and $x \in pa_{L_*}(c) \setminus P \cup \{t\}$ in case that **IV.** holds but **III.** does not hold. Lemma 4.2 then concludes the proof. □

## 7  Conclusion

Let us summarize the results and conjectures. The following conditions on DAGs $K$ and $L$ over $N$ were shown to be equivalent in case $|\mathcal{E}(K)| \leq |\mathcal{E}(L)|+1$: the inclusion $\mathcal{I}(K) \subseteq \mathcal{I}(L)$, the inclusion conditions for $K$ in $L$, the graphical conditions (a)-(e) and the existence of a sequence of DAGs $L = G_1, \ldots, G_n = K, n \geq 1$ in which each next DAG is obtained by legal arrow reversal or adding. We conjecture that the described ideas can be extended to a general case. Confirmation of our conjectures could have positive impact on the methods of learning Bayesian networks.

### Acknowledgements

This research was supported by the grants GAČR n. 201/01/1482, FRVŠ n. 2001/1433 and GAAVČR n. K1019101. The authors benefited from participation in the seminar "Conditional independence structures" (Toronto, October 1999) and HSSS research kitchen "Learning conditional independence models" (Třešť, October 2000). We thank the anonymous reviewers for useful remarks.